\documentclass[conference,letterpaper]{IEEEtran}
\IEEEoverridecommandlockouts
\usepackage[utf8]{inputenc}
\usepackage{cite}
\usepackage{amsmath,amssymb,amsfonts}
\usepackage{algorithmic}
\usepackage{graphicx}
\usepackage{textcomp}
\usepackage{xcolor}
\usepackage{makecell}
\usepackage{url}
\def\BibTeX{{\rm B\kern-.05em{\sc i\kern-.025em b}\kern-.08em
    T\kern-.1667em\lower.7ex\hbox{E}\kern-.125emX}}
\begin{document}

\title{Improvements to short-term weather prediction with recurrent-convolutional networks 
\thanks{This project benefited from parallel development in the fellowship ``Seamless Artificially Intelligent Thunderstorm Nowcasts'' from the European Organisation for the Exploitation of Meteorological Satellites (EUMETSAT).}}

\author{\IEEEauthorblockN{1\textsuperscript{st} Jussi Leinonen}
\IEEEauthorblockA{
\textit{Federal Office of Meteorology and Climatology MeteoSwiss}\\
Locarno-Monti, Switzerland \\
jussi.leinonen@meteoswiss.ch, ORCID 0000-0002-6560-6316}
}

\maketitle

\begin{abstract}
The Weather4cast 2021 competition gave the participants a task of predicting the time evolution of two-dimensional fields of satellite-based meteorological data. This paper describes the author's efforts, after initial success in the first stage of the competition, to improve the model further in the second stage. The improvements consisted of a shallower model variant that is competitive against the deeper version, adoption of the AdaBelief optimizer, improved handling of one of the predicted variables where the training set was found not to represent the validation set well, and ensembling multiple models to improve the results further. The largest quantitative improvements to the competition metrics can be attributed to the increased amount of training data available in the second stage of the competition, followed by the effects of model ensembling. Qualitative results show that the model can predict the time evolution of the fields, including the motion of the fields over time, starting with sharp predictions for the immediate future and blurring of the outputs in later frames to account for the increased uncertainty.
\end{abstract}

\begin{IEEEkeywords}
weather forecasting, neural networks, gated recurrent units, optimizers, model ensembling
\end{IEEEkeywords}

\section{Introduction}

Neural networks have recently been investigated for many areas of weather forecasting, including nowcasting, data assimilation, acceleration of the evaluation of model physics, and the postprocessing of model outputs (e.g. \cite{Ravuri2021GenerativePrecipitation,Geer2021AssimilationML,Seifert2020MicrophysicsML,Haupt2021MLPostprocessing}), with ambitious plans in place to further expand their use in the future \cite{Dueben2021ECMWFRoadmap}. As a complex spatiotemporal problem, weather forecasting benefits from the ability of convolutional networks to learn spatial relationships and the ability of recurrent networks to learn the rules of temporal evolution. Therefore, improvements to the architecture, training and application of these networks can benefit several different applications within the area of weather forecasting.

The Weather4cast 2021\footnote{\url{https://www.iarai.ac.at/weather4cast/}} Stage 1 challenge\cite{Gruca2021Weather4cast} was organized to encourage the development of neural networks for the weather forecasting field. The challenge gave the competitors a task of developing machine learning models that predict the future evolution of satellite-derived weather data fields given several time steps of the past state of those fields. The four variables considered were \textit{temperature} (either cloud top temperature when cloudy, or the surface temperature otherwise), convective rainfall rate (\textit{crr\_intensity}), probability of tropopause folding (\textit{asii\_turb\_trop\_prob}), and cloud mask (\textit{cma}). The challenge was divided into two parts: in the ``Core'' competition, the networks were evaluated in regions where they were also trained, while in the ``Transfer Leaning'' competition, the evaluation was performed in regions where training data were not available, emphasizing the ability of the model to generalize outside the training data. The competition entry by the present author \cite{Leinonen2021Weather4castStage1} used a recurrent-convolutional neural network with modified Convolutional Gated Recurrent Unit (ConvGRU, \cite{Ballas2016ConvGRU}) layers organized in a multiscale encoder-forecaster architecture. This architecture proved to be an effective solution to the problem posed in the challenge and achieved first place in both the Core and Transfer Learning competitions.

After the conclusion of Stage 1, Weather4cast 2021 continued with the IEEE Big Data Cup competition. This competition was identical in structure to the original except for the addition of new data regions for both training and evaluation. Therefore, improvements could be made by making optimal use of the new data and by improving upon the models used in Stage 1. This paper presents the improvements made by the author in the IEEE Big Data Cup competition that resulted in better model performance. The code that can be used to replicate the results can be found in \url{https://github.com/jleinonen/weather4cast-bigdata}; the weights for the pre-trained models are provided at \cite{Leinonen2021WeightsBigData}.

\section{Models and training} \label{sect:models}

The predictions posted by the author to the IEEE Big Data Cup competition utilized the same basic model design used in the Stage 1 competition and described in \cite{Leinonen2021Weather4castStage1}. The IEEE Big Data Cup competition dataset included more training data in the form of new regions. These data were utilized by retraining the models with the new dataset; furthermore, the static data provided by the competition organizers, consisting of surface elevation, latitude and longitude, were used for each of the regions. No external data sources were used, nor were any calculations performed to generate additional input data, except for data augmentation with mirroring and rotation. Modifications were also made to the models and the training procedure to further improve the performance. These included a shallower variant of the model, a change to a different optimizer, conditional modeling to address representativeness issues with \textit{crr\_intensity}, and model ensembling to reduce the loss. The sections below describe the improvements in more detail. Other than these improvements, the models and the training procedure were the same as in \cite{Leinonen2021Weather4castStage1}, and the reader is referred to that paper for the details.

\subsection{Shallower model variant} \label{sect:shallow}

In the Stage 1 competition, two variants of the recurrent-convolutional model were investigated: one that used standard ConvGRU gates and another (ResGRU) that utilized residual units in place of the convolutions. In this work, the model selection was expanded with a variant where the deepest layer of the original ResGRU model was removed. This variant was found to outperform the deeper network in some cases, and had the further advantage of using much fewer weights ($4.6$ million, compared to $18.6$ million in the full-size model).

\subsection{AdaBelief optimizer}

While the original worked used the widely adopted Adam optimizer, in this work AdaBelief \cite{Zhuang2020AdaBelief}, a modification of the algorithm used in Adam \cite{Kingma2014Adam}, was adopted. According to its authors, AdaBelief improves the generalization ability of the solutions it finds compared to those found by Adam. Further information can be found in the repository at \url{https://github.com/juntang-zhuang/Adabelief-Optimizer}.

The hyperparameters of AdaBelief and comparisons to Adam were examined using the \textit{temperature} variable. After some experimentation with the parameters, AdaBelief with the default hyperparameters was adopted, with the $\epsilon$ parameter set to $10^{-14}$ (the default of the TensorFlow implementation), as non-default parameters either degraded the results or failed to provide a significant benefit. The runs with variables other than \textit{temperature} were performed with AdaBelief using these hyperparameters, as the time remaining in the contest did not allow for parameter tuning for each variable separately.

\subsection{Adjustments to rain rate prediction} \label{sect:crr}

Predicting \textit{crr\_intensity} involves particular challenges that do not apply to the other variables because often it is not raining at all, while much of the rainfall is delivered by only a few events. In Stage 1, it was already noted that a different model performed best for \textit{crr\_intensity} compared to the other variables. In further investigation of the statistics of this variable, it was also found that there were issues with the representativeness of the training dataset: The validation set had an average \textit{crr\_intensity} of $7.26 \times 10^{-4}$ on the $[0,1]$ normalized scale, while for the training set it was considerably lower, $6.01 \times 10^{-4}$. The equivalent discrepancies for the other variables were at least an order of magnitude smaller.

The problem of different distributions in the training and validation datasets was approached by building a conditional model containing two copies of the same model. The results from one of these copies would be used if the maximum rain rate in the input time series was less than a threshold value of $0.026$ on the $[0,1]$ normalized scale; otherwise the other copy would be used. The threshold is was chosen such that it divides the training dataset approximately evenly over the two models. The conditional approach helps adjust for the representativeness problem because, as the validation set contains more raining cases than the training set, the model used for the higher rain rates would be used more often to compensate for this.

Two such conditional models were trained for \textit{crr\_intensity}. In the first, the conditional model contained two copies of the ConvGRU architecture used for \textit{crr\_intensity} in \cite{Leinonen2021Weather4castStage1}; before training, the weights of both copies were initialized to a model trained for the entire dataset. In the second, slightly differently constructed ConvGRU models were used to avoid certain inconvenient architectural features of the older model, and the entire conditional model was trained from random initialization.

\subsection{Model ensembling} \label{sect:ensembling}

The experimentation with different types of models and hyperparameters produced multiple relatively good models. This presented an opportunity to further improve the results by ensembling the models. As most of the experimentation and tuning was done with the \textit{temperature} variable, this variable had more ensemble members available. For \textit{temperature}, the ensemble consisted of variants of the shallow model (see Sect.~\ref{sect:shallow}); the deep model weights were not used as they became incompatible with the final model architecture due to changes during the development process. For \textit{asii\_turb\_trop\_prob} and \textit{cma}, the ensemble consisted of a mixture of deep and shallow models, while the \textit{crr\_intensity} ensemble used the two different conditional models described in Sect.~\ref{sect:crr}. The final ensembles had five members for \textit{temperature}, three for \textit{asii\_turb\_trop\_prob} and two each for \textit{crr\_intensity} and \textit{cma}. 

Three ensembling strategies were attempted to produce predictions for the final leaderboard. In the first, members were equally weighted with the weights summing to $1$. In the second, optimal weights $\mathbf{w}$ were derived using ridge regression \cite{Hoerl1970Ridge}, also called Tikhonov regularization \cite{Boyd2004Convex}, which gives weights that optimize mean square error as
\begin{equation}
\mathbf{w} = \left( \mathbf{X}^\mathrm{T}\mathbf{X} + \lambda\mathbf{I} \right)^{-1} \mathbf{X}^\mathrm{T}\mathbf{y} \label{eq:weights}
\end{equation}
where $\mathbf{X}$ is a dimension $(n,p)$ matrix with $p$ models and $n$ predictions per model, $\mathbf{y}$ is a $(n,1)$ vector with the correct values, and $\lambda$ is a regularization parameter. For all variables, $\lambda$ was set to $10^{-4}$ times the mean of the diagonal of $\mathbf{X}^\mathrm{T}\mathbf{X}$. The matrices $\mathbf{X}^\mathrm{T}\mathbf{X}$ and $\mathbf{X}^\mathrm{T}\mathbf{y}$ were computed using the validation dataset due to suspicion that slight overfitting of the models to the training set might weaken the derivation of the weights. This also motivated a third strategy as a variant of the second, as it was noted that the weights for \textit{crr\_intensity} summed to a total of $1.0526$, even as the sum of weights for the other three variables deviated from $1$ by less than $0.004$. This likely reflects the representativeness issue that was already noted in Sect.~\ref{sect:crr}. While the nonzero weighting considerably improved the metric for the validation set, it was not clear if the systematic difference would also be valid for the held-out dataset. Therefore, a solution was also created where the weights for \textit{crr\_intensity} were derived as optimal weights with the sum constrained to $1$. These can be derived using the method of Lagrange multipliers \cite{Boyd2004Convex}, which gives
\begin{equation}
\begin{bmatrix} \mathbf{w} \\ \mu \end{bmatrix} =
\begin{bmatrix}
\mathbf{X}^\mathrm{T}\mathbf{X} & -\frac{1}{2}\mathbf{q} \\ 
\mathbf{q}^\mathrm{T} & 0 \\ 
\end{bmatrix}^{-1}
\begin{bmatrix}
\mathbf{X}^\mathrm{T}\mathbf{y} \\ 1
\end{bmatrix} \label{eq:weights-constrained}
\end{equation}
where $\mu$ is a constraint parameter that can be discarded after solving the equation, and $\mathbf{q} = [1, 1, ..., 1]^\mathrm{T}$ is a $(p,1)$ vector with all elements set to $1$.

The ensemble weights are shown in Table~\ref{table:weights}. The constrained optimal weights produced by Eq.~\ref{eq:weights-constrained} are also shown for all variables for reference purposes. However, no submission was attempted with these other than for \textit{crr\_intensity}, as for the other variables the weights given by Eq.~\ref{eq:weights} were already close to $1$ and the negative weight given by Eq.~\ref{eq:weights-constrained} for one of the \textit{temperature} ensemble member was deemed suspicious.

\begin{table}[tbp]
\caption{Ensemble weights}
\begin{center}
\begin{tabular}{lcccc}
\hline
& \textit{temperature} & \textit{crr\_intensity} & \textit{asii\_turb\_} & \textit{cma} \\
& & & \textit{trop\_prob} & \\
\hline\hline
\textbf{Ridge} & & & & \\
(1) & $0.1455$ & $0.5206$ & $0.4344$ & $0.4864$ \\
(2) & $0.2666$ & $0.5320$ & $0.2722$ & $0.5165$ \\
(3) & $0.0904$ & & $0.2941$ & \\
(4) & $0.2487$ & & & \\
(5) & $0.2457$ & & & \\
\hline
\textbf{Constrained} & & & & \\
(1) & $0.2094$ & $0.5122$ & $0.4592$ & $0.4846$ \\
(2) & $0.3224$ & $0.4878$ & $0.2613$ & $0.5154$ \\
(3) & $-0.0664$ & & $0.2795$ & \\
(4) & $0.2660$ & & & \\
(5) & $0.2687$ & & & \\
\end{tabular}
\label{table:weights}
\end{center}
\end{table}

\section{Results}

In the course of model development, the various model candidates were evaluated using the validation dataset after training and the results were recorded. A summary of these is shown in Table~\ref{table:metrics}. The first row shows the results using the best models from Stage 1 without retraining for the new dataset. Models D1 and D2 are different training runs of the deeper variant, models S1--S5 use the shallow variant, while C1 and C2 are the conditional models used for \textit{crr\_intensity}. The results for the equally weighted ensemble and the ensemble with weights obtained using ridge regression are also shown. It should be noted that models were trained for each variable separately; for example, the S1 model for \textit{temperature} and the S1 model for \textit{cma} only share a common architecture, not common weights. The different versions of the shallow model for \textit{temperature} are as follows: S1 was trained from random initialization using the default settings of the AdaBelief optimizer, S2 was trained by initializing the weights with those of S1 and resetting the learning rate to $10^{-3}$ in the beginning of training, S3 was trained like S2 but setting the $\epsilon$ parameter to $10^{-7}$ (another value suggested by the authors), S4 was trained like S1 but enabling weight decay in AdaBelief, and S5 was trained like S1 but using the Adam optimizer instead.
\begin{table}[tbp]
\caption{Summary of the validation set metrics of the different variables and models}
\begin{center}
\begin{tabular}{lllll}
\hline
& \textit{temperature} & \textit{crr\_intensity} & \makecell[l]{\textit{asii\_turb\_}\\\textit{trop\_prob}} & \textit{cma} \\
\hline\hline
Stage 1 & $0.004539$ & $0.00009126$ & $0.002165$ & $0.1512$ \\
\hline
\textbf{Deep} & & & & \\
D1 & $0.004370$ & $0.00009107$ & $0.002033^{\mathrm{1}}$ & $0.1440^{\mathrm{1}}$ \\
D2 & $0.004335$ & $0.00009122$ & $0.002021^{\mathrm{2}}$ &  \\
\hline
\textbf{Shallow} & & & & \\
S1 & $0.004345^{\mathrm{1}}$ & & $0.002065^{\mathrm{3}}$ & $0.1438^{\mathrm{2}}$ \\
S2 & $0.004349^{\mathrm{2}}$ & & &  \\
S3 & $0.004344^{\mathrm{3}}$ & & &  \\
S4 & $0.004355^{\mathrm{4}}$ & & &  \\
S5 & $0.004378^{\mathrm{5}}$ & & &  \\
\hline
\textbf{Cond.} & & & & \\
C1 & & $0.00008960^{\mathrm{1}}$ & &  \\
C2 & & $0.00008968^{\mathrm{2}}$ & &  \\
\hline
\textbf{Ensemble} & & & & \\
Equal & $0.004285$ & $0.00008872$ & $0.001989$ & $0.1423$ \\
Ridge & $0.004282$ & $0.00008588$ & $0.001987$ & $0.1424$ \\
\hline
\multicolumn{5}{l}{$^{\mathrm{1-5}}$Included in the ensembles (Table~\ref{table:weights})} 
\end{tabular}
\label{table:metrics}
\end{center}
\end{table}

Multiple general trends can be identified from the results of Table~\ref{table:metrics}. First, it can be seen that simply retraining the model with the new data and the AdaBelief optimizer improved the results considerably (up to $4.8\%$ for \textit{cma}) over the models that produces the best results in Stage 1, except for \textit{crr\_intensity} where the improvement was more marginal ($0.2\%$). Between the retraining and the new optimizer, the addition of new training data seems to be the more important factor, as the S5 model trained with Adam was only slightly worse than the other models trained with AdaBelief. Second, the adoption of the conditional model for \textit{crr\_intensity} provided an improvement ($1.8\%$) for that variable that was more significant than the retraining. Third, the ensembling further improves the scores for all variables. The ensemble of equally weighted models provided benefits of $1.0\%$--$1.6\%$ over the best individual model, depending on the variables. Ridge regression weighting provided results that were very similar to the equal-weights ensemble for all variables except for \textit{crr\_intensity}, where the improvement was large ($3.2\%$) compared to the other benefits from ensembling. This is likely because the ensemble weights for \textit{crr\_intensity}, whose sum differs considerably from $1$, correct for the bias between the training and validation sets notes in Sect.~\ref{sect:ensembling}. However, it was not clear \textit{a priori} if this bias would also apply to the test and held-out datasets.

\begin{figure*}[tbp]
\centerline{\includegraphics[width=\linewidth]{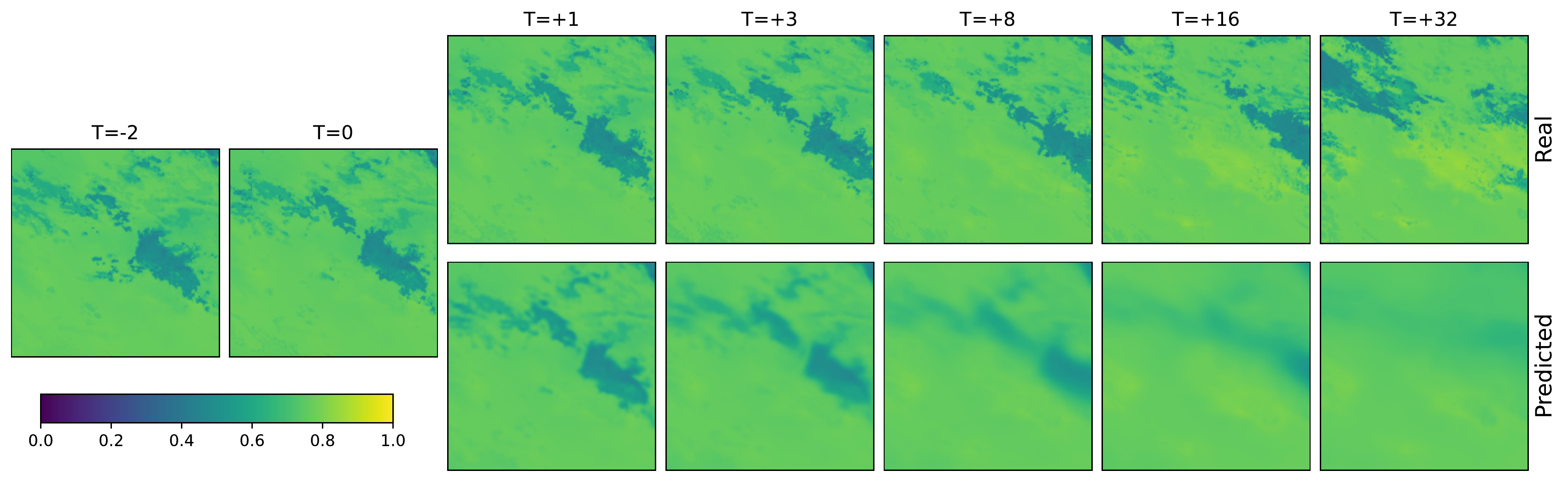}}
\caption{An example of a model prediction for the \textit{temperature} variable using the ridge regression model ensemble. The left side shows the past temperature, while the right side show the real temperature in the future (top row) and the corresponding model prediction (bottom row). $T$ is the index of the frame in the sequence; $T=0$ refers to the last input frame and $T=1$ is the first prediction. Scaled temperature normalized to the range $(0,1)$, covering the data range in the whole dataset, is shown.}
\label{fig:prediction_temp}
\end{figure*}

\begin{figure*}[tbp]
\centerline{\includegraphics[width=\linewidth]{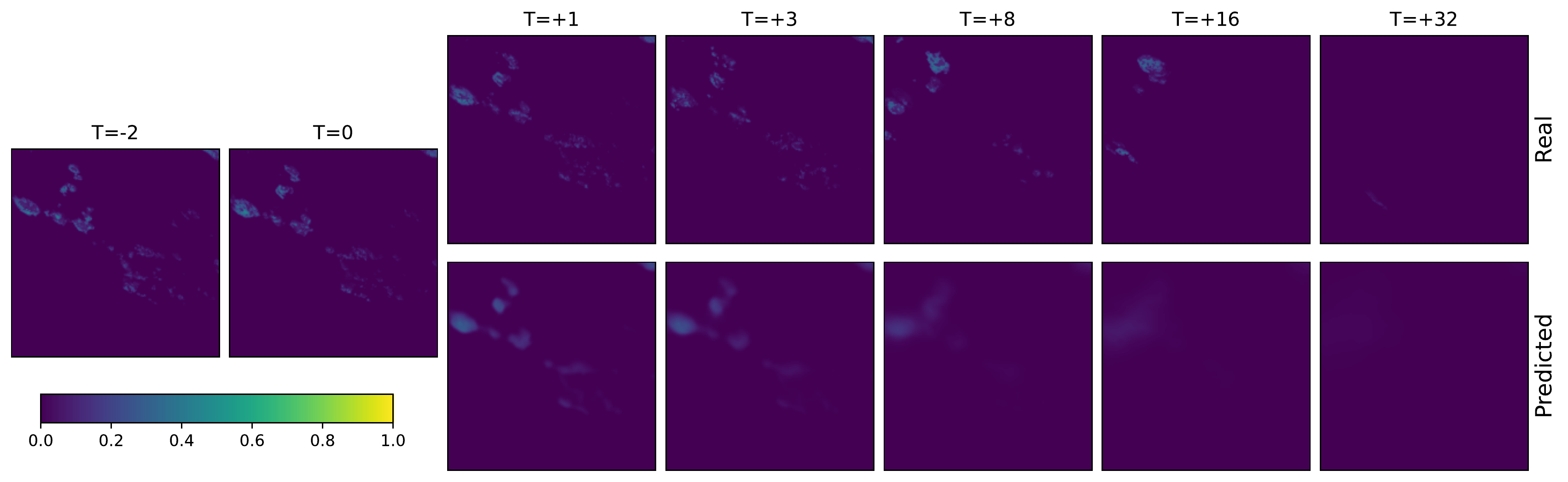}}
\caption{As Fig.~\ref{fig:prediction_temp}, but for \textit{crr\_intensity}. A different case is shown because the case in Fig.~\ref{fig:prediction_temp} contains negligible precipitation.}
\label{fig:prediction_crr}
\end{figure*}

\begin{figure*}[tbp]
\centerline{\includegraphics[width=\linewidth]{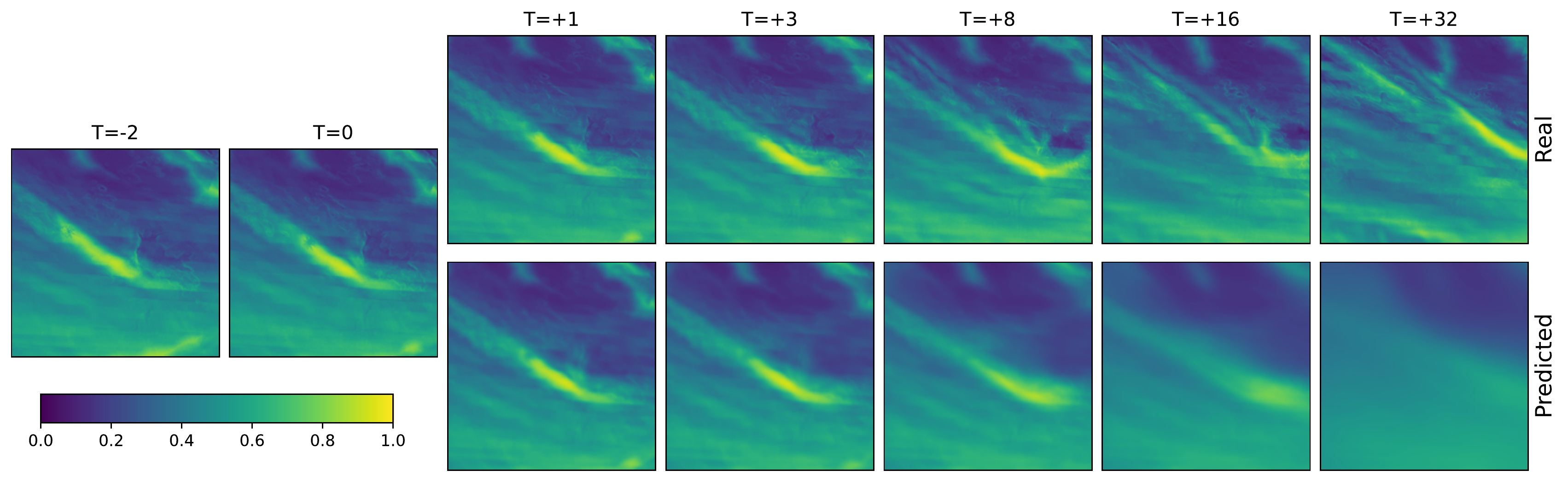}}
\caption{As Fig.~\ref{fig:prediction_temp}, but for \textit{asii\_turb\_trop\_prob}.}
\label{fig:prediction_attp}
\end{figure*}

\begin{figure*}[tbp]
\centerline{\includegraphics[width=\linewidth]{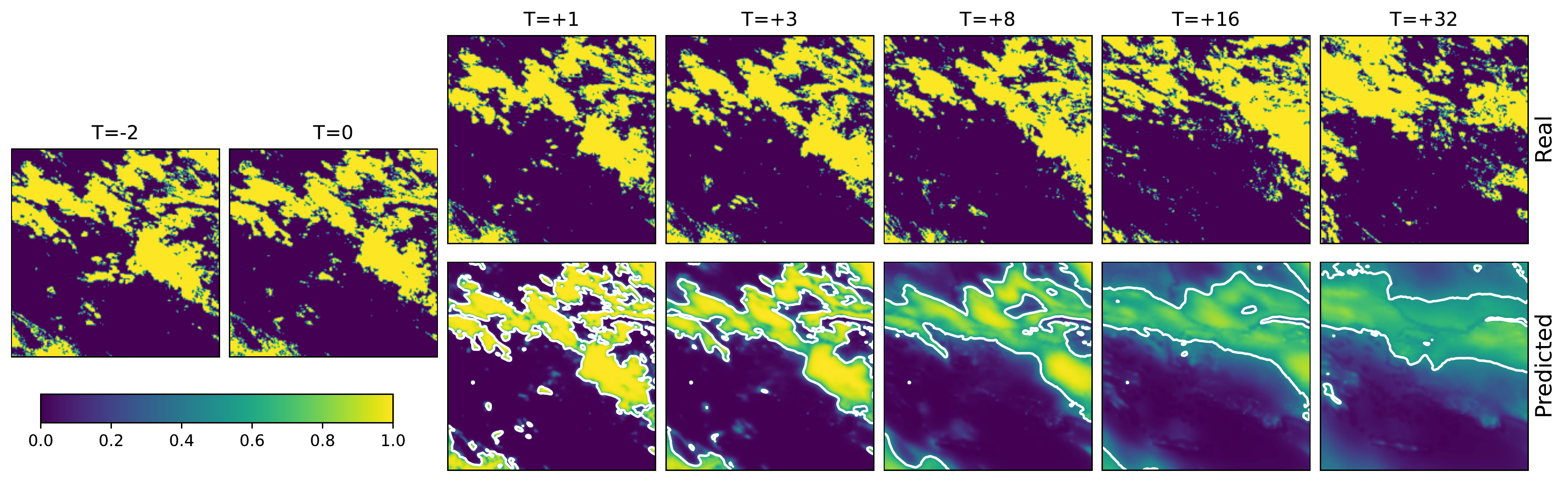}}
\caption{As Fig.~\ref{fig:prediction_temp}, but for \textit{cma}. The white lines show the contour at $0.5$, which is the decision threshold for a binary cloud mask.}
\label{fig:prediction_cma}
\end{figure*}

Figs.~\ref{fig:prediction_temp}--\ref{fig:prediction_cma} show the results of prediction with the ensemble models using the ridge regression weights. Like the predictions in \cite{Leinonen2021Weather4castStage1}, these show that the model is able to detect and predict the motion of the input fields. Predictions start sharp and become blurrier with longer lead times, reflecting the increasing uncertainty of the prediction.

The results achieved on the competition leaderboards are shown in Table~\ref{table:leaderboards}. The first two columns show the test leaderboard scores for the Core and Transfer Learning competitions, respectively, while the third and fourth column show the equivalent scores for the final leaderboards which used the held-out dataset and determined the results of the competition. The retraining improved the scores considerably, as the best single models outperform the Stage 1 models by $3.0\%$ for the Core competition and $5.4\%$ for the Transfer Learning competition. The conditional model for \textit{crr\_intensity} improved the scores slightly, and the ensembling provided a benefit comparable to the improvements seen in Table~\ref{table:metrics}. Using optimized weights from Eq.~\ref{eq:weights} improved the scores on the final leaderboards, but only marginally. 

\begin{table}[tbp]
\caption{Results on the leaderboards}
\begin{center}
\begin{tabular}{lcccc}
\hline
& \textbf{Test} & \textbf{Test} & \textbf{Final} & \textbf{Final} \\
& \textbf{Core} & \textbf{Transfer} & \textbf{Core} & \textbf{Transfer} \\
\hline\hline
Stage 1 & $0.4757$ & $0.4552$ & & \\[1mm]
\makecell[l]{Best\\Single} & $0.4615$ & $0.4308$ & & \\[2mm]
\makecell[l]{Conditional\\rain} & $0.4605$ & $0.4289$ & & \\[2mm]
\makecell[l]{Equal\\Ensemble} & $\mathbf{0.4530}$ & $\mathbf{0.4229}$ & $0.4734$ & $0.4328$ \\[2mm]
\makecell[l]{Ridge\\Ensemble} & & & $\mathbf{0.4729}$ & $\mathbf{0.4323}$ \\
\hline
\end{tabular}
\label{table:leaderboards}
\end{center}
\end{table}

\section{Conclusions}
The results of the efforts to improve the model offer various insights into the training of multi-scale recurrent-convolutional models for image time series prediction tasks and particularly weather forecasting:
\begin{enumerate}
    \item The models appear to be able to learn better from a larger amount of training data; this was most clearly evidenced by the fact that the scores in the Transfer Learning competition increased significantly, indeed more than those in the Core competition, after retraining with additional data. This indicates that the additional data improved the ability of the model to generalize. While the larger effect in the Transfer Learning competition might also have resulted from the new Core dataset being more representative of the Transfer Learning regions, it is not clear if this is the case.
    \item The model introduced in \cite{Leinonen2021Weather4castStage1} performs well also in a variant that has the deepest layers removed. This model attained similar results to the original, deeper variant, sometimes outperforming it and sometimes being slightly worse. The shallower variant has much fewer weights than the deep variant, although its computational and memory requirements are not much smaller as these are determined mostly by the top levels of the network.
    \item The AdaBelief optimizer seems to somewhat outperform the popular Adam optimizer at this task, giving better results with the validation dataset. The results obtained in this study support the claim by the authors of AdaBelief that results obtained with it generalize better to datasets outside the training data.
    \item For datasets where the training data do not represent well the validation and/or testing data, such as \textit{crr\_intensity} in this competition, using conditional models (Sect. \ref{sect:crr}) can improve the results. However, the solution adopted here --- choosing the model based on the maximum rain rate in the input sequence --- was rather \textit{ad hoc}, and better can likely be found. Indeed, better solutions are likely already documented in the literature, but the time constraints of the competition prohibited an exhaustive search for the best method.
    \item Ensembling provided an improvement on the order of $1\%$ to the scores. While such an advantage can potentially be decisive in a competition environment, it depends on the use case whether such an improvement is worth the added training in real-world applications. However, researchers who test different models and hyperparameters during the development process may benefit from saving the different variants and finally ensembling them, as this provides the ensemble members without additional resources spent on training.
\end{enumerate}

Among potential further improvements, the author recommends investigating the role of regularization, normalization and dropout, which were only examined in a very limited fashion in this study, as they did not produce obvious benefits in the model investigated here, but were very successfully used by other challenge participants (e.g. \cite{Kwok2021Variational}) to limit overfitting. Another interesting technique to investigate would be transformer networks, which have proven very effective for modeling sequences; another competitor used transformers, and while the model presented here outperformed the transformer model in the competition, it may be possible to find a way to combine the advantages of the two approaches. Additional input variables could also further improve the results; while the author did not find substantial benefits from using inputs beyond the four target variables (possibly because the experimentation was performed with the \textit{temperature} variable), the other competitors' models did benefit from them. Furthermore, the prediction of the \textit{cma} variable, which is essentially a binary classification problem, could benefit from an alternative loss function. These were attempted in this study but not studied thoroughly, so further research is needed.

\section*{Acknowledgment}
The author thanks U. Hamann and A. Rigazzi for discussions regarding the model and training.

\bibliographystyle{IEEEtranDOI}
\bibliography{weather4cast_bigdata}

\end{document}